%% file: main.tex
\documentclass[final]{cvpr}

\usepackage{times}
\usepackage{epsfig}
\usepackage{graphicx}
\usepackage{amsmath}
\usepackage{amssymb}
\usepackage{arydshln}
\usepackage{booktabs}

\usepackage{enumitem}
\usepackage[dvipsnames]{xcolor}
\usepackage{comment}
\usepackage{textcomp}

\definecolor{amethyst}{rgb}{0.6, 0.4, 0.8}

\input{mathcommands}

\usepackage[colorlinks,citecolor=teal]{hyperref}

\begin{document}

\title{Information-Theoretic Segmentation by Inpainting Error Maximization}

\author{
\begin{tabular}[t]{c@{\extracolsep{4em}}c} 
Pedro Savarese  & Sunnie S. Y. Kim \\
TTI-Chicago & Princeton University \\ 
{\tt\small savarese@ttic.edu} & {\tt\small suhk@cs.princeton.edu}
\end{tabular}
\\~\\
\begin{tabular}[t]{c@{\extracolsep{4em}}c@{\extracolsep{4em}}c}
Michael Maire  & Greg Shakhnarovich & David McAllester \\
University of Chicago & TTI-Chicago & TTI-Chicago \\
{\tt\small mmaire@uchicago.edu} & {\tt\small greg@ttic.edu} & {\tt\small mcallester@ttic.edu}
\end{tabular}
}

\maketitle

\begin{abstract}
   We study image segmentation from an information-theoretic perspective, proposing a novel adversarial method that performs unsupervised segmentation by partitioning images into maximally independent sets. More specifically, we group image pixels into foreground and background, with the goal of minimizing predictability of one set from the other. An easily computed loss drives a greedy search process to maximize inpainting error over these partitions. Our method does not involve training deep networks, is computationally cheap, class-agnostic, and even applicable in isolation to a single unlabeled image. Experiments demonstrate that it achieves a new state-of-the-art in unsupervised segmentation quality, while being substantially faster and more general than competing approaches.\footnote{Code is available at \url{https://github.com/lolemacs/iem}}
  
\end{abstract}

\input{introduction}

\input{relatedwork}

\input{method}

\input{experiments}

\input{discussion}

{\small
\bibliographystyle{ieee_fullname}
\bibliography{egbib}
}

\clearpage
\input{appendix}

\end{document}

%% file: mathcommands.tex
\renewcommand{\eqref}[1]{(\ref{#1})}

\newcommand{\dist}{\mathcal D}

\usepackage{amsmath,amsfonts,bm}

\def\eqref#1{equation~\ref{#1}}
\def\Eqref#1{Equation~\ref{#1}}

\def\1{\bm{1}}

\DeclareMathAlphabet{\mathsfit}{\encodingdefault}{\sfdefault}{m}{sl}
\SetMathAlphabet{\mathsfit}{bold}{\encodingdefault}{\sfdefault}{bx}{n}

\newcommand{\R}{\mathbb{R}}



%% file: introduction.tex
\section{Introduction}

Deep neural networks have significantly advanced a wide range of computer
vision capabilities, including
image classification~\cite{alexnet,vgg,googlenet,resnet1},
object detection~\cite{rcnn,yolo,retinanet},
and semantic segmentation~\cite{deeplab,pspnet}.
Nonetheless, neural networks typically require massive amounts of manually
labeled training data to achieve state-of-the-art performance.  Applicability
to problems in which labeled data is scarce or expensive to obtain often
depends upon the ability to transfer learned representations from related
domains.

These limitations have sparked exploration of self-supervised methods for
representation learning, where an automatically-derived proxy task guides deep
network training.  Subsequent supervised fine-tuning on a small labeled dataset
adapts the network to the actual task of interest.  A common approach to
defining proxy tasks involves predicting one part of the data from another,
\eg, geometric relationships~\cite{doersch2015unsupervised,
noroozi2016unsupervised,gidaris2018unsupervised},
colorization~\cite{larsson2017colorization},
inpainting~\cite{pathakCVPR16context}.
A recent series of advances focuses on learning representations through a
contrastive objective~\cite{learning_noise,unsup_inst,cpc,tian2019contrastive},
and efficiency scaling such systems~\cite{moco,simclr,byol} to achieve parity
with supervised pre-training.

\begin{figure}[!tb]
    \centering
    \begin{minipage}[t]{1.00\linewidth}
    \begin{minipage}[t]{0.19\linewidth}
      \vspace{3pt}
      \centering
      \scriptsize\textbf{\textsf{Image}}
    \end{minipage}
    \hfill
    \begin{minipage}[t]{0.19\linewidth}
      \vspace{3pt}
      \centering
      \scriptsize\textbf{\textsf{Ground-truth}}
    \end{minipage}
    \hfill
    \begin{minipage}[t]{0.19\linewidth}
      \vspace{3pt}
      \centering
      \scriptsize\textbf{\textsf{IEM Result}}
    \end{minipage}
    \hfill
    \begin{minipage}[t]{0.19\linewidth}
      \vspace{0pt}
      \centering
      \scriptsize\textbf{\textsf{Inpainted Foreground}}
    \end{minipage}
    \hfill
    \begin{minipage}[t]{0.19\linewidth}
      \vspace{0pt}
      \centering
      \scriptsize\textbf{\textsf{Inpainted Background}}
    \end{minipage}
    \end{minipage}
    \begin{minipage}[t]{1.00\linewidth}
    \begin{minipage}[t]{0.2\linewidth}
      \vspace{0pt}
      \centering
      \includegraphics[width=\linewidth]{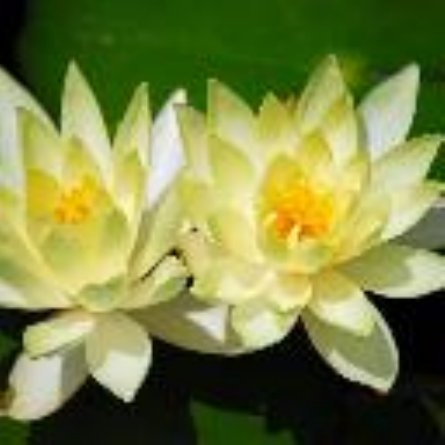}\\\vspace{-1pt}%
      \includegraphics[width=\linewidth]{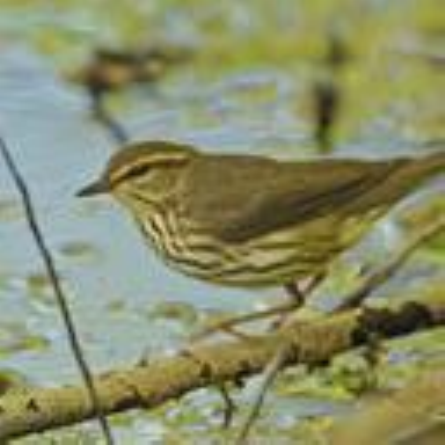}\\\vspace{-1pt}%
      \includegraphics[width=\linewidth]{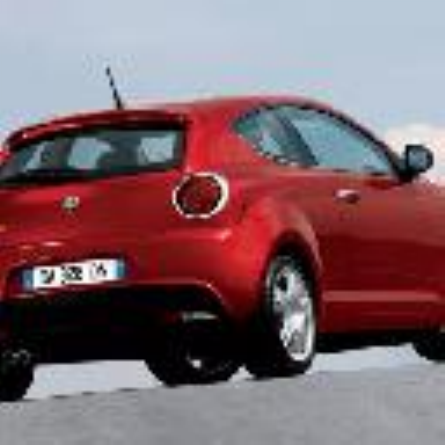}\\\vspace{-1pt}%
    \end{minipage}
    \hspace{-5pt}
    \begin{minipage}[t]{0.2\linewidth}
      \vspace{0pt}
      \centering
      \includegraphics[width=\linewidth]{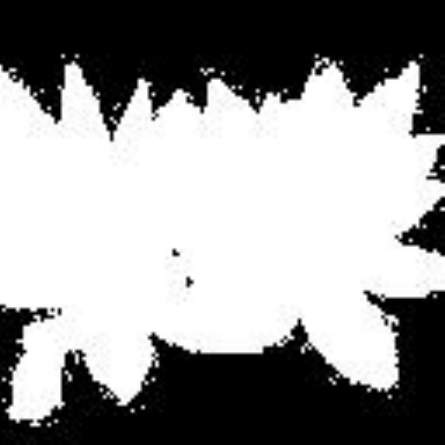}\\\vspace{-1pt}%
      \includegraphics[width=\linewidth]{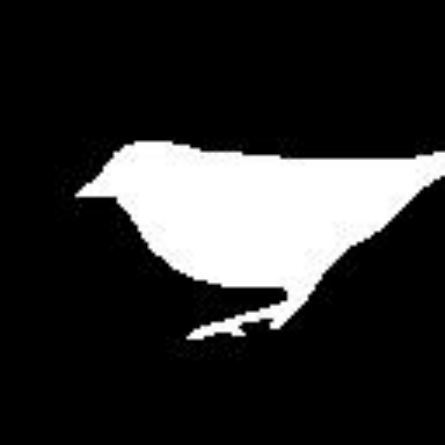}\\\vspace{-1pt}%
      \includegraphics[width=\linewidth]{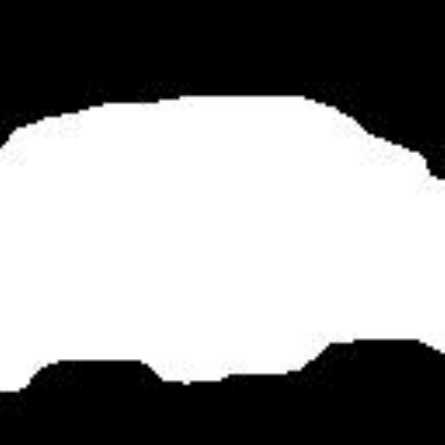}\\\vspace{-1pt}%
    \end{minipage}
    \hspace{-5pt}
    \begin{minipage}[t]{0.2\linewidth}
      \vspace{0pt}
      \centering
      \includegraphics[width=\linewidth]{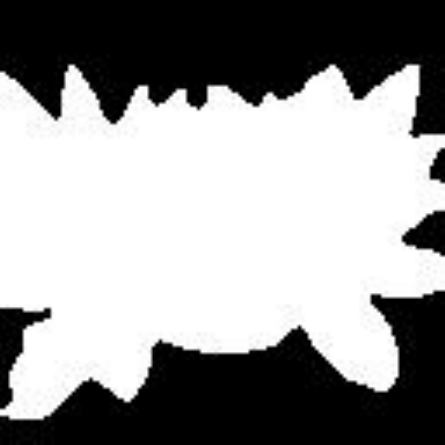}\\\vspace{-1pt}%
      \includegraphics[width=\linewidth]{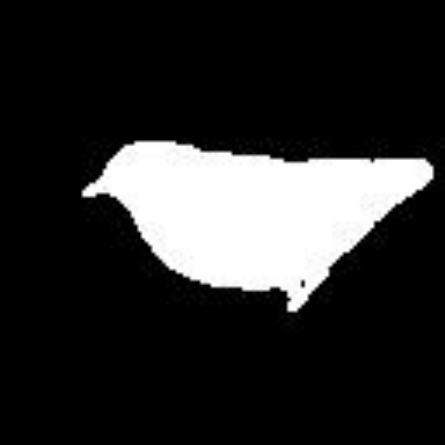}\\\vspace{-1pt}%
      \includegraphics[width=\linewidth]{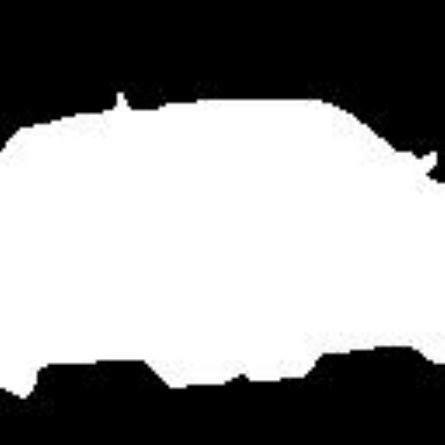}\\\vspace{-1pt}%
    \end{minipage}
    \hspace{-5pt}
    \begin{minipage}[t]{0.2\linewidth}
      \vspace{0pt}
      \centering
      \includegraphics[width=\linewidth]{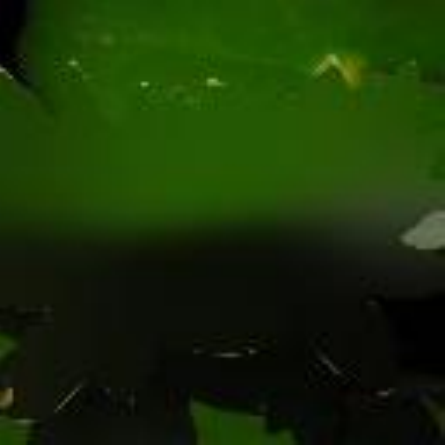}\\\vspace{-1pt}%
      \includegraphics[width=\linewidth]{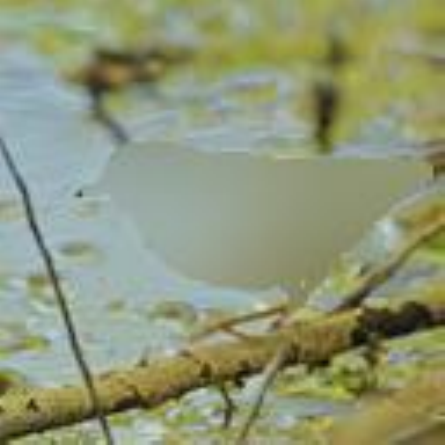}\\\vspace{-1pt}%
      \includegraphics[width=\linewidth]{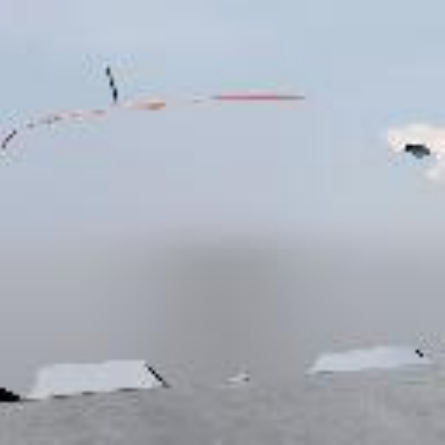}\\\vspace{-1pt}%
    \end{minipage}
    \hspace{-5pt}
    \begin{minipage}[t]{0.2\linewidth}
      \vspace{0pt}
      \centering
      \includegraphics[width=\linewidth]{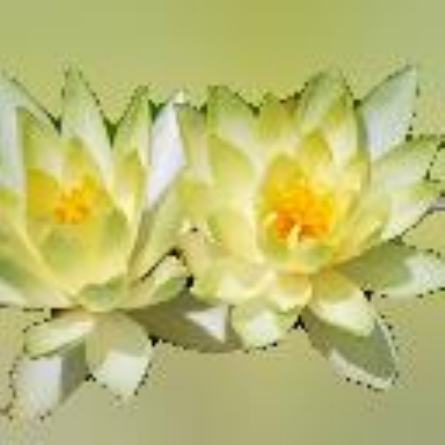}\\\vspace{-1pt}%
      \includegraphics[width=\linewidth]{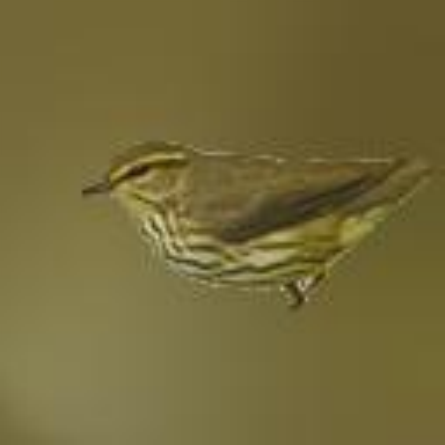}\\\vspace{-1pt}%
      \includegraphics[width=\linewidth]{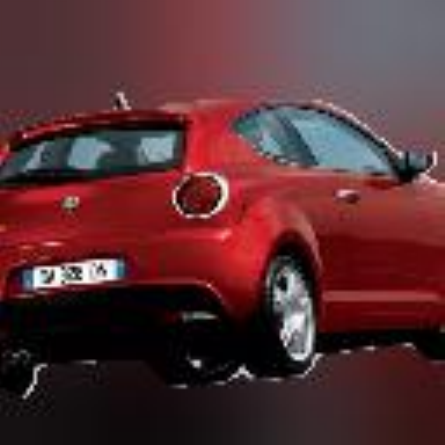}\\\vspace{-1pt}%
    \end{minipage}
    \end{minipage}
    \vspace{2pt}
    \caption{Illustration of our Inpainting Error Maximization (IEM) framework for
    completely unsupervised segmentation, applied to flowers, birds, and cars.
    Segmentation masks maximize the error of inpainting foreground given background
    and vice-versa.}
    \label{fig:intro_examples}
\end{figure}

Another class of approaches frames unsupervised learning within a generative
modeling context, building upon, \eg, generative adversarial networks
(GANs)~\cite{gan} or variational autoencoders (VAEs)~\cite{vae}.
Donahue~\etal~\cite{bigan,bigbigan} formulate representation learning using a
bidirectional GAN.  Deep InfoMax~\cite{deepinfomax} drives unsupervised
learning by maximizing mutual information between encoder inputs and outputs.
InfoGAN~\cite{infogan}, which adds a mutual information maximization objective
to a GAN, demonstrates that deep networks can learn to perform image
classification without any supervision---at least for small-scale datasets.

Inspired by this latter result, we focus on the question of whether more
complex tasks, such as image segmentation, can be solved in a purely
unsupervised fashion, without reliance on any labeled data for training or
fine-tuning.  We address the classic task of generic, category-agnostic
segmentation, which aims to partition any image into meaningful regions
(\eg, foreground and background), without relying on knowledge of a
predefined set of object classes.

\begin{figure*}[!ht]
    \centering
    \includegraphics[width=\linewidth]{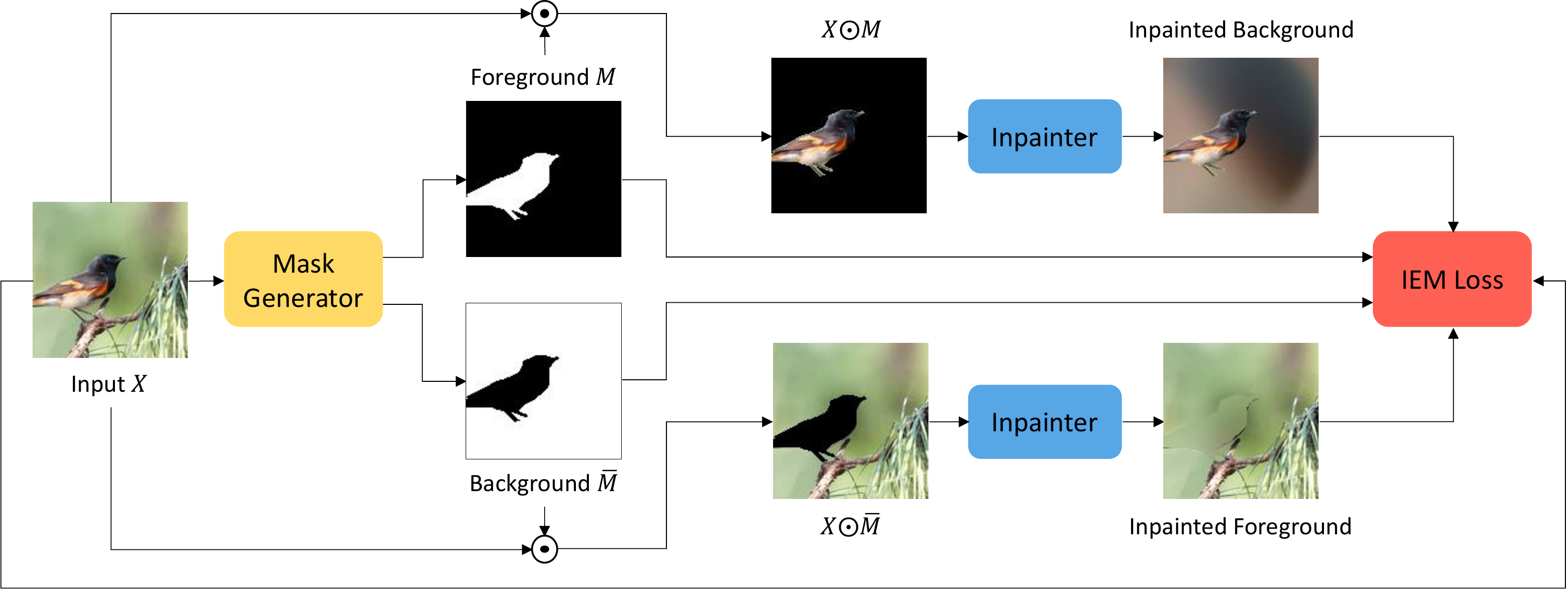}
    \caption{%
    Inpainting Error Maximization (IEM) framework.
    Given an unlabeled image $X$, a mask generator module first produces
    segmentation masks (\eg, foreground $M$ and background $\bar{M}$).  Each
    mask selects a subset of pixels from the original image by performing an
    element-wise product between the mask and the image, hence partitioning
    the image into regions. Inpainting modules try to reconstruct each
    region given all others in the partition, and the IEM loss is defined by a
    weighted sum of inpainting errors.}
   \label{fig:framework}
\end{figure*}

Here we introduce \emph{Inpainting Error Maximization (IEM)} as an approach to unsupervised segmentation. IEM is motivated
by the intuition that a segmentation into objects minimizes the mutual information between the pixels in the segments, and hence makes inpainting of one segment given the others difficult.  This gives a natural adversarial objective where a segmenter tries to maxmize, while an inpainter tries to minimize, inpainting error.
However, rather than adopt an adversarial training objective we find it more effective to fix a basic inpainter and directly maximize inpainting error through a form of gradient descent on the segmentation. Our version of IEM is learning-free and can be applied directly to any image in any domain. Figure~\ref{fig:intro_examples} shows example results for
foreground-background segmentation derived from our IEM method which is diagrammed in Figure~\ref{fig:framework}.

We show that the segmentations produced by the learning-free
IEM segmenter 
can be used as noisy training labels to train a deep segmentation
network which further improves our segmentation quality.
This bootstrapping does not utilize human generated labels
and our system has no equivalent of fine-tuning.

While IEM has a natural adversarial nature, we avoid employing a GAN.  This contrasts with recent GAN-based
unsupervised segmentation approaches, such as ReDO~\cite{REDO_NeurIPS2019}
and PerturbGAN~\cite{PerturbedGAN_NeurIPS2019}, which Section~\ref{sec:related}
reviews in detail.  Experiments in Section~\ref{sec:experiments} demonstrate
that our learning-free method matches or outperforms both. Our work
advances unsupervised segmentation via the following contributions:
\begin{itemize}[topsep=0pt,noitemsep]
    \item{%
        An information-theoretic inspired IEM procedure
        for image segmentation which is fast, learning-free, and can be applied directly to any image in any domain.%
    }%
    \vspace{2pt}
    \item{%
        Extensive empirical results showing that our IEM procedure performs competitively with
        prior work on unsupervised segmentation when measured in terms of
        intersection-over-union (IoU).
    }%
    \vspace{2pt}
    \item{%
        An optional refinement phase for IEM wherein a neural
        segmentation network is trained on a subset of images and their IEM segmentations and where the training images are selected to be those having high IEM inpainting error. This network can then be incorporated into the IEM process, resulting in a system that comfortably
        outperforms all competing methods.
    }%
\end{itemize}

Our results put dominant approaches to unsupervised
segmentation into question.  In comparison to IEM, generative modelling not
only results in more computationally expensive methods, but also fails at
learning high-quality segmentations. IEM provides a new combination of modeling and learning, and perhaps a new direction for unsupervised methods.

%% file: relatedwork.tex
\section{Related Work}
\label{sec:related}

Semantic segmentation, as a pixel-level category labeling problem, has rapidly
advanced due to availability of large-scale labeled datasets and supervised
deep learning~\cite{FCN,U-Net,badrinarayanan2015segnet,zhao2017pspnet,
DeebLabV3,Chen_2018_ECCV}.  These tools have yielded similarly impressive
progress on segmentation of detected object instances~\cite{rcnn,
he2017maskrcnn}.  In absence of large-scale annotation, the now common
self-supervised approach, consisting of proxy task driven representation
learning followed by fine-tuning, has demonstrated successful transfer to
semantic and object segmentation tasks~\cite{moco}.  Various efforts have also
explored weak supervision from
image labels~\cite{pathakICCV15ccnn,Papandreou_2015_ICCV,Huang_2018_CVPR},
bounding boxes~\cite{Xia_2013_ICCV,Dai_2015_ICCV,Papandreou_2015_ICCV,
Khoreva_2017_CVPR},
or saliency maps~\cite{joon17cvpr,Zeng_2019_ICCV,Wang_2018_CVPR}.

Less clear is the suitability of these weakly- or self-supervised approaches
for category-agnostic image segmentation, which could be regarded as primarily
a grouping or clustering problem, as opposed to a labeling task.  Indeed,
class-agnostic segmentation has inspired a diverse array of algorithmic
approaches.  One framework that has proven amenable to supervised deep learning
is that of considering the dual problem of
edge detection~\cite{deepedge,deepcontour,hed,kokkinos2015pushing}
and utilizing classic clustering algorithms and morphological transformations
to convert edges into regions~\cite{ncut,gpb-ucm}.

Unsupervised class-agnostic segmentation methods instead often directly
address the partitioning task.  Earlier work clusters pixels using color,
contrast, and hand-crafted features~\cite{czmhh_contrastSaliency_cvpr11,
Jiang_2013_CVPR,Zhu_2014_CVPR}.  More recent strategies include
self-training a convolutional neural network~\cite{kanezaki2018}, and
maximizing mutual information between cluster assignments~\cite{
Ji_2019_ICCV}.

A particular line of recent unsupervised work uses generative models to
partition an image.  Chen~\etal~\cite{REDO_NeurIPS2019} propose a GAN-based
object segmentation model, ReDO (ReDrawing of Objects), premised on the idea
that it should be possible to change the appearance of objects without
affecting the realism of the image containing them.  During training, the
model's mask generator infers multiple masks for a given image, and the region
generator ``redraws'' or generates new pixel values for each mask's region,
one at a time.  The discriminator then judges the newly composed image for
realism.  After training, passing a novel image through the learned mask
generator suffices to segment it.

Bielski \& Favaro~\cite{PerturbedGAN_NeurIPS2019} propose a different
GAN-based model, which we refer to as PerturbGAN.  They build on the idea
that one can perturb object location without affecting image realism.
Comprising their model is an encoder that maps an image into a latent code,
a generator that constructs a layered representation consisting of a
background image, a foreground image, and a mask, and a discriminator that
assesses the layered representation.  During training, small random shifts
applied to the foreground object assist with learning.  Inference proceeds
by encoding the image and feeding the latent code to the trained generator.

In a different direction, Voynov~\etal~\cite{voynov2020big} examine the latent
space of an off-the-shelf GAN and obtain saliency masks of synthetic images
via latent space manipulations.  Using these masks, they train a segmentation
model with supervision.  On another front,
Benny \& Wolf~\cite{OneGAN_ECCV2020} train a complex model (OneGAN), with
multiple encoders, generators, and discriminators, to solve several tasks
simultaneously, including foreground segmentation.  Their model is
weakly-supervised, requiring class labels and clean background images but
not pixel- or region-level annotation.

As modern techniques addressing the fully unsupervised setting,
ReDO~\cite{REDO_NeurIPS2019} and PerturbGAN~\cite{PerturbedGAN_NeurIPS2019}
serve as a reference for experimental comparison to our IEM method.  We do
not compare to the methods of Voynov~\etal~\cite{voynov2020big} or Benny \&
Wolf~\cite{OneGAN_ECCV2020}, as the former involves manual examination of
latent space manipulation directions and the latter utilizes weak supervision
from additional datasets.

Our methodology relates to other recent information-theoretic
segmentation approaches, though our setting and framework differ. Yang~\etal~\cite{cis, yang2020timesupervised} segment objects in video by minimizing
the mutual information between motion field partitions, which they approximate with an adversarial inpainter. We likewise focus on inpainting objectives, but in a manner not anchored to trained adversaries and not reliant on video dynamics.

Wolf~\etal~\cite{Wolf2020BMVC} segment cells in microscopy images by
minimizing a measure of information gain between partitions.  They approximate
this information gain measure using an inpainting network, and segment
images of densely populated cells in a hierarchical fashion.  While
philosophically aligned in terms of objective, our optimization strategy
and algorithm differs from that of Wolf~\etal.  We search for a partition
over the global image context, whereas Wolf~\etal focus on partitioning
local image patches in a manner reminiscent of edge detection.

%% file: method.tex
\section{Information-theoretic Segmentation}
\label{sec:method}

We model an image as a composition of independent layers, \eg, foreground and background, and aim to recover these layers by partitioning so
as to maximize the error of inpainting one partition from the others.
Our full method utilizes Inpainting Error Maximization (IEM) as the primary component in the first of a two-phase procedure.  In this first phase, a greedy binary optimization algorithm produces, independently for each image,
a segmentation mask assigning each pixel to a partition, with the goal to maximize mutual inpainting error. We  approximate the inpainting error by a simple filtering procedure yielding an objective that is easy and cheap to compute.

We emphasize that there is no \emph{learning} in this initial phase, meaning that segmentation masks produced for each image are independent of any other images, resulting in a computationally cheap and distribution-agnostic subroutine---\ie it can be applied even to a single unlabeled image and, unlike prior work, does not require segmented objects to be of similar semantic classes.

When we do have a collection of images sharing semantics, we can apply the second phase, wherein we select masks from the first phase with the highest inpainting error (which we deem the most successful) and use them as label supervision to train a segmentation network. This phase refines suboptimal masks from the first phase, while also enforcing semantic consistency between segmentations. 
Note that our second phase is optional, and although we observe empirical improvements from training a neural segmentation model, the quality of masks produced by optimizing IEM alone are comparable and often superior to ones learned by competing methods.

\newcommand{\img}{X\xspace}
\newcommand{\imgspace}{\mathcal{X}\xspace}
\newcommand{\bimg}{B_{|M}}
\newcommand{\fimg}{F_{|M}}

\subsection{Segmenting by Maximizing Inpainting Error}
Consider an image $\img \in \imgspace = \R^{C \times H \times W}$ with $C$ channels, height $H$ and width $W$. We assume that $\img$ is generated by a stochastic process: first, foreground and background images $F \in \R^{C \times H \times W}$ and $B \in \R^{C \times H \times W}$ are drawn independently from distributions $\dist_F$ and $\dist_B$, respectively. Next, a binary segmentation mask $M(F,B) \in \mathcal M = \{0,1\}^{1 \times H \times W}$ is deterministically produced by $F$ and $B$. Finally, the image is generated by composing the foreground and background as $\img = F \odot M + B \odot \overline M = \fimg + \bimg$, where $\overline M = 1 - M, \fimg = F \odot M, \bimg = B \odot \overline M$, and $\odot$ denotes the Hadamard product.

Moreover, we assume that the mapping $M$ is injective and can be equivalently represented as $M = M(\img)$. That is, the mask can be fully recovered from the image $\img$.

Our framework relies on information-theoretic measures, particularly the mutual information between two random variables $A$ and $Z$:
\begin{equation}
    I(A,Z) = H(A) - H(A|Z) = H(Z) - H(Z|A) \,,
\end{equation}
where $H(A)$ is the entropy of $A$ and $H(A|Z)$ is the conditional entropy of $A$ given $Z$.

Mutual information between two random variables is zero if and only if they are independent. Under our assumption that $F$ and $B$ are independent, it follows that $I(F,B) = 0$. By the data-processing inequality of mutual information, it also follows that $I(\fimg,\bimg) = 0$.

\begin{figure}[!t]
    \centering
    \includegraphics[width=\linewidth]{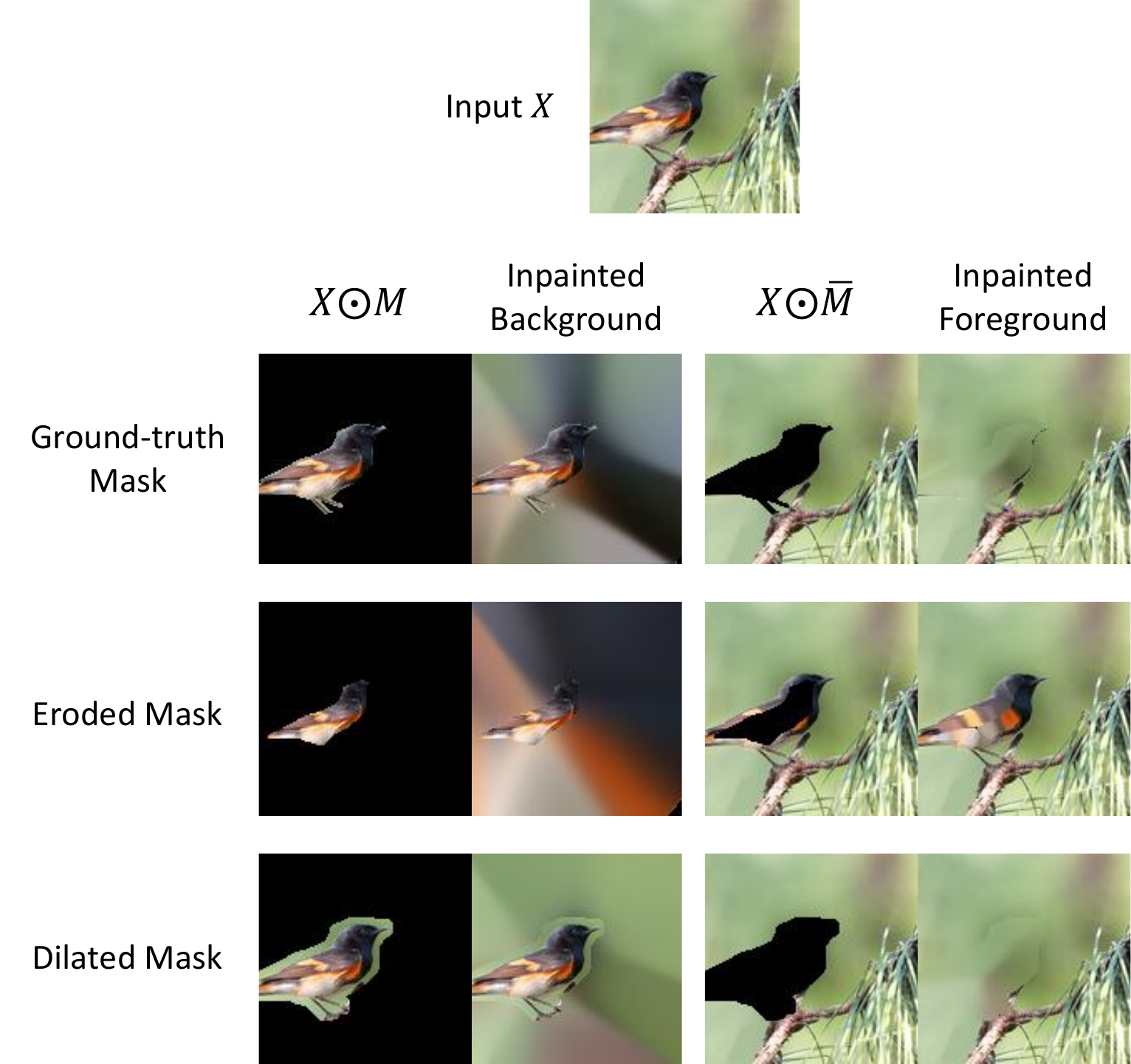}
    \caption{Foreground and background inpainting results using ground-truth, eroded (smaller), and dilated (bigger) masks. We see that the ground-truth mask incurs high inpainting error for both the foreground and the background, while the eroded mask allows reasonable inpainting of the foreground and the dilated mask allows reasonable inpainting of the background. Hence, we expect IEM, which maximizes the inpainting error of each partition given the others, to yield a segmentation mask close to the ground-truth.}\label{fig:gtmask}
\end{figure}

We consider the task of partitioning each image $\img$ into two sets of pixels $\hat F$ and $\hat B$ such that $I(\hat F, \hat B) = 0$. The hope is that $\hat F = \fimg$ and $\hat B = \bimg$, hence fully recovering the ground-truth segmentation. However, this problem admits a trivial solution where either $\hat F$ or $\hat B$ is empty, in which case we have $I(\hat F, \hat B) = 0$, a minimizer.
We circumvent this issue by using a normalized variant of the mutual information, namely the \textit{coefficient of constraint} \cite{mathpsychology, Lancichinetti_2009, recipes}:
\begin{equation}
    \begin{split}
    C(A,Z) &= \frac{I(A,Z)}{H(A)} + \frac{I(Z,A)}{H(Z)} \\
        & = 2 - \left( \frac{H(A|Z)}{H(A)} + \frac{H(Z|A)}{H(Z)}  \right) \,.
    \end{split}
\end{equation}

Similar to the mutual information, we have that $C(A,Z) = 0$ if and only if $A$ and $Z$ are independent. On the other hand, we have that one of the denominators approaches $0$ as either $A$ or $Z$ approaches the empty set. Therefore, partitioning each $\img$ into $\hat F$ and $\hat B$ to minimize $C(\hat F, \hat B)$, as an alternative to minimizing $I(\hat F, \hat B)$, partially avoids the trivial solution where either $\hat F$ or $\hat B$ is empty.

Now, let $\phi : \imgspace \to \mathcal M$ denote an arbitrary mapping from images to binary segmentation masks, and define random variables $\hat F_\phi = \img\odot \phi(\img)$ and $\hat B_\phi = \img \odot \overline{ \phi(\img)}$, which are regions of the image partitioned by $\phi$. Our goal is then to find $\phi$ that minimizes the coefficient of constraint between the predicted foreground and background:
\begin{equation}
    \min_\phi \,\, C(\hat F_\phi, \hat B_\phi) = \max_\phi \,\, \frac{H(\hat F_\phi| \hat B_\phi)}{H(\hat F_\phi)} + \frac{H(\hat B_\phi | \hat F_\phi)}{H(\hat B_\phi)}.
    \label{eq-mincoef}
\end{equation}

This problem provides a general framework for unsupervised segmentation, where the distributions over $\hat F$ and $\hat B$ required to compute $C(\hat F_\phi, \hat B_\phi)$ can be arbitrarily chosen.

We model the conditional probabilities required to approximate $H(\hat F_\phi| \hat B_\phi)$ and $H(\hat B_\phi | \hat F_\phi)$ as $\ell_1$-Laplacians with identity covariances and conditional means $\psi$, yielding expectations over $\|\hat F_\phi - \psi(\hat B_\phi)\|_1$ and $\|\hat B_\phi - \psi(\hat F_\phi)\|_1$.

For the marginals that define $H(\hat F_\phi)$ and $H(\hat B_\phi)$, we opt for an agnostic approach which avoids introducing any bias on foreground and background pixel values. More specifically, we assign uniform distributions over pixel values, which results in expectations over $\|\phi(\img)\|$ and $\|\overline{\phi(\img)}\|$.

Appendix~\ref{app:iem-details} discusses our choices of distributions in more detail, including how they lead to the approximations above. We also empirically compare our approximations against other options in Appendix~\ref{app:ablation-norm}, adopting the same experimental protocol which is formalized later in Section~\ref{sec:experiments}.

Since the model $\psi$ of the conditional means should be chosen such that the probability of observed foreground and background regions is maximized (\ie the respective entropies are minimized), we must also account for optimization over $\psi$, which adds an adversarial component to the optimization problem in \Eqref{eq-mincoef}:
\begin{equation}
    \begin{split}
    \max_\phi \min_\psi\,\, & \frac{\mathbb E \left[\|\img \odot \phi(\img) - \psi(\img \odot \overline{\phi(\img)}) \|_1 \right]}{\mathbb E \left[\|\phi(\img) \| \right]}\\
    & + \frac{\mathbb E \left[\|\img \odot \overline{\phi(\img)} - \psi(\img \odot \phi(\img) \|_1 \right]}{\mathbb E \left[\| \overline{\phi(\img)} \| \right]} \,.
    \end{split}
    \label{eq-iem}
\end{equation}

Note that if the marginal and conditional probabilities required to characterize $C(\hat F_\phi, \hat B_\phi)$ in Equation~\ref{eq-mincoef} were instead chosen to be Gaussians, the objective would be equivalent to the Contextual Information Separation (CIS) criterion~\cite{cis,yang2020timesupervised} applied to raw images instead of motion fields. The objective above can thus be seen as a variant of CIS with different density models to approximate $C(\hat F_\phi, \hat B_\phi)$.

Objectives defined by an $\ell_1$ inpainting error normalized by the mask size (\emph{normalized inpainting error}), as in \Eqref{eq-iem}, have also been employed to train inpainters in, for example, Yu~\etal~\cite{inpaintingattention,inpaintinggated}\footnote{See version 1.0.0 of the official repository.} and Lin~\cite{gatedpytorch}, giving another interpretation to the objective above: a maximization of the normalized inpainting error.

To illustrate the idea, we show qualitative results of inpainted foreground and background using ground-truth, eroded (smaller), and dilated (bigger) masks in Figure~\ref{fig:gtmask}. We see that the ground-truth mask incurs high inpainting error for both the foreground and the background, while the eroded mask allows reasonable inpainting of the foreground and the dilated mask allows reasonable inpainting of the background. 
Hence we expect IEM, which maximizes the inpainting error of each partition given the others, to drive a predicted mask closer to the ground-truth.

\subsection{Fast \& Distribution-agnostic Segmentation}
\label{sec-distributionagnostic}

We design a procedure to approximately solve the IEM objective in Equation~\ref{eq-iem} given a \emph{single} unlabeled image $\img$.

We start by introducing additional assumptions regarding our adopted image generation process: we assume that pixels of $F$ and $B$ have strong spatial correlation and hence can be approximately predicted by a weighted average of their neighbors. Let $K$ denote the kernel of a Gaussian filter with standard deviation $\sigma$ and arbitrary size, \ie
\begin{equation}
    K_{i,j} \propto \frac{1}{2 \pi \sigma^2} \exp \left(- \frac{(i-\mu_i)^2 + (j-\mu_j)^2}{2 \sigma^2} \right) \,,
\end{equation}
where $\mu_i, \mu_j$ are the center pixel positions of the kernel, and $K$ is normalized such that its elements sum to one. Then, we adopt the following inpainting module:
\begin{equation}
    \psi_K(X, M) = \frac{K * X}{K * M} \,.
    \label{eq-inpainter}
\end{equation}

Lastly, if the feasible set of $\phi$ is arbitrarily expressive, \eg, the set of all functions from $\mathcal X$ to $\mathcal M$, then optimizing $\phi$ for a single image is equivalent to directly searching for a binary mask $M \in \mathcal M = \{0,1\}^{1 \times H \times W}$:
\begin{equation}
    \begin{split}
    \max_{M \in \mathcal M} \quad & \frac{\| M \odot( \img - \psi_K(\img \odot \overline M, \overline M)) \|_1}{\|M\|}\\
    & + \frac{\|\overline M \odot (\img - \psi_K(\img \odot M, M)) \|_1}{\|\overline{M}\|}
    \end{split}
    \label{eq-iem-direct}
\end{equation}

Let $\mathcal L_{inp}$ denote the above objective as a function of $M$, for a fixed image $X$. We optimize $M$ with projected gradient ascent, adopting the update rule:
\begin{equation}
    M_{t+1} = \mathcal P_{S(M_t)} \Big(M_t + \eta \nabla_M \mathcal L_{inp}(M_t) \Big) \,,
\end{equation}
where $\eta$ is the learning rate, $\mathcal P$ is the projection operator, and $S(M_t) \subseteq \mathcal M$ is a local feasible set that depends on $M_t$.

Our preliminary experiments show that taking $\eta \to \infty$ results in faster increases of the inpainting objective, while also simplifying the method since, in this case, there is no need to tune the learning rate. The updates become:
\begin{equation}
    M_{t+1} = \mathcal P^\infty_{S(M_t)} \Big(\nabla_M \mathcal L_{inp}(M_t) \Big) \,,
\end{equation}
where $\mathcal P^\infty(M) = \lim_{\eta \to \infty} \mathcal P(\eta M)$, to be detailed below.

We define $S(M)$ as the set of binary masks that differ from $M$ only in its \emph{boundary pixels}. More specifically, we say that a pixel is in the boundary of $M$ if it contains a pixel in its 4-neighborhood with the opposite (binary) value. Therefore, the projection $\mathcal P_{S(M)}$ has two roles: first, it restricts updates to pixels in the boundary of $M$ (all other pixels of $M$ have their values unchanged during updates); second, it ensures that new masks are necessarily binary by projecting each component of its argument to the closest value in $\{0,1\}$. It then follows that $\mathcal P^\infty_{S(M)}$ projects negative components to 0 and positive components to 1.

We also propose a regularization mechanism that penalizes the diversity in each of the extracted image layers. Let $\sigma(M)$ denote the \emph{total} deviation of pixels in $X \odot M$:
\begin{equation}
    \sigma(M) = \|M \odot (X - \mu(M)) \|_2^2\,, \label{eq-reg}
\end{equation}
where $\mu(M)$ is the average pixel value of $X \odot M$ (not counting masked-out pixels). We define the regularized IEM objective as $\mathcal L_{IEM}(M) = \mathcal L_{inp}(M) - \frac{\lambda}{2} \left(\sigma(M) + \sigma(\overline M) \right)$, where $\lambda$ is a non-negative scalar that controls the strength of the diversity regularizer.

Finally, we adopt a simple smoothing procedure after each mask update to promote more uniform segmentation masks. For each pixel in $M$, we compute the sum of pixel values in its 8-neighborhood, and assign a value of 1 to the pixel in case the sum exceeds 4 and a value of 0 otherwise.

\subsection{Mask Refinement by Iterative Learning}

In this optional refinement phase, applicable when we have a \emph{dataset} of unlabeled images known to share some semantics (\eg, same object category forming the foreground) we select masks produced by IEM with highest inpainting error (which we deem the most successful) and use them as labels to train a neural segmentation model with supervision. The goal is to improve mask quality and promote semantic consistency, as IEM performs distribution-agnostic segmentation. After the model is trained, we can, again optionally, return to the first phase and further refine the mask inferred by the segmentation model with IEM.

%% file: experiments.tex
\section{Experiments}
\label{sec:experiments}

We first describe our experimental setup and evaluation metrics. We then discuss qualitative results of IEM.
Finally, we compare IEM quantitatively to recently proposed unsupervised segmentation methods, ReDO and PerturbGAN, and to the classic GrabCut algorithm.

\subsection{Datasets}

We demonstrate IEM on the following datasets, and compare its performance to that of prior work where possible.
\textbf{Caltech-UCSD Birds-200-2011 (CUB)}~\cite{CUB-200-2011} consists of 11,788 images of 200 classes of birds and segmentation masks.
\textbf{Flowers}~\cite{Flowers} consists of 8,189 images of 102 classes of flowers, with segmentation masks obtained by an automated algorithm developed specifically for segmenting flowers in color photographs~\cite{Flowers_Segmentation}.
\textbf{LSUN Car}~\cite{LSUN}, as part of the large-scale LSUN dataset of 10 scene categories and 20 objects, consists of 5,520,753 car images. Segmentation masks are not provided, so following Bielski \& Favaro~\cite{PerturbedGAN_NeurIPS2019}, we approximate ground-truth masks for the first 10,000 images with Mask R-CNN~\cite{he2017maskrcnn}, pre-trained on the COCO~\cite{COCO} dataset with a ResNet-50 FPN backend. We used the pre-trained model from the Detectron2 library~\cite{wu2019detectron2}. Cars were detected in 9,121 images, and if several cars were detected, we grabbed the mask of the biggest instance as ground truth.

\begin{table}[!tb]
\setlength{\tabcolsep}{7pt}
\caption{Unsupervised segmentation results on Flowers, measured in terms of accuracy, IoU, and DICE score. Segmentation masks used for evaluation are publicly available ground-truth.} \label{tab:flowers}
\centering
\vspace{1pt}
\begin{tabular}{@{}lrrr@{}}
\toprule
                                    & Accuracy  & IoU  & DICE  \\ \cmidrule{2-4}
GrabCut \cite{GrabCut}              & 82.0      & 69.2 & 79.1  \\ 
ReDO \cite{REDO_NeurIPS2019}        & 87.9      & 76.4 & ---   \\ \hdashline
IEM (ours)                          & 88.3      & 76.8 & 84.6  \\ 
IEM+SegNet (ours)                   & \textbf{89.6} & \textbf{78.9} & \textbf{86.0}  \\ \bottomrule
\end{tabular}
\end{table}

\begin{table}[!tb]
\setlength{\tabcolsep}{7pt}
\caption{Unsupervised segmentation results on CUB, measured in terms of accuracy, IoU, and DICE score. Segmentation masks used for evaluation are publicly available ground-truth.} \label{tab:birds}
\centering
\vspace{1pt}
\begin{tabular}{@{}lrrr@{}}
\toprule
                                            & Accuracy  & IoU   & DICE  \\ \cmidrule{2-4}
GrabCut \cite{GrabCut}                      & 72.3      & 36.0  & 48.7  \\ 
PerturbGAN \cite{PerturbedGAN_NeurIPS2019}  & ---       & 38.0  & ---   \\ 
ReDO \cite{REDO_NeurIPS2019}                & 84.5      & 42.6  & ---   \\ \hdashline
IEM (ours)                                  & 88.6      & 52.2  & 66.0  \\ 
IEM+SegNet (ours)                           & \textbf{89.3} & \textbf{55.1} & \textbf{68.7}  \\ \bottomrule
\end{tabular}
\end{table}

\begin{table}[!tb]
\setlength{\tabcolsep}{7pt}
\caption{Unsupervised segmentation results on LSUN Car, measured in terms of accuracy, IoU, and DICE score. Segmentation masks used for evaluation were automatically generated with Mask R-CNN, following PerturbGAN~\cite{PerturbedGAN_NeurIPS2019}.} \label{tab:cars}
\centering
\vspace{1pt}
\begin{tabular}{@{}lrrr@{}}
\toprule
                                            & Accuracy  & IoU   & DICE  \\ \cmidrule{2-4}
GrabCut \cite{GrabCut}                      & 69.1      & 57.6  & 71.8  \\ 
PerturbGAN \cite{PerturbedGAN_NeurIPS2019}  & ---       & 54.0  & ---   \\ \hdashline
IEM (ours)                                  & 76.2      & 65.1  & 78.1  \\ 
IEM+SegNet (ours)                           & \textbf{77.8} & \textbf{68.5} & \textbf{80.5}  \\ \bottomrule
\end{tabular}
\end{table}

\subsection{Implementation Details}

For all experiments, we adopt the same training, validation, and test splits as ReDO for CUB and Flowers (resulting in 1,000 and 1,020 test images, respectively), and random subsampling of 1,040 test images for LSUN Car. Preliminary experiments to guide the design of our method were performed on the validation set of CUB.

We run IEM for a total of 150 iterations on the test set of each dataset, with an inpainter $\psi_K$ whose convolutional kernel $K$ is of size $21 \times 21$ and has a scale $\sigma = 5$, which, for computational reasons, we approximate by two stacked convolutions with kernel sizes $11 \times 11$. Moreover, we set the strength of the diversity regularizer as $\lambda = 0.001$.

All images are resized and center-cropped to $128 \times 128$ pixels. Running all 150 iterations of IEM on a set of $\approx$1000 images takes under 2 minutes on a single Nvidia 1080 Ti.

Masks are initialized with centered squares of varying sizes. For each dataset we run IEM, adopting squares with size 44, 78, and 92 as initialization, and only consider the results whose initialization lead to the highest inpainting error. Hence, ground truth labels are not used to choose mask initializations, and there is no feedback between test evaluation and the size of squares used to initialize the masks.

For the optional refinement phase, we select the 8,000 masks produced by IEM that induce the highest inpainting error and train a variant of PSPNet~\cite{zhao2017pspnet} to perform segmentation using the collected masks as pseudo-labels. Following ReDO, our model consists of an initial resolution-decreasing residual block, followed by three resolution-preserving residual blocks and pyramid pooling, where all batch norm layers~\cite{bn} are replaced by instance norm~\cite{in}.

We train our segmentation model for a total of 50 epochs to minimize the pixel-wise binary cross-entropy loss. The network is optimized with Adam~\cite{adam}, a constant learning rate of $10^{-3}$, and a mini-batch size of 128.

\subsection{Evaluation Metrics}

In our framework, the predicted and ground-truth masks are binary and have 1 in foreground pixels and 0 in background pixels. We evaluate the predicted masks' quality with three commonly used metrics. First, we measure the (per-pixel) mean \emph{accuracy} of the foreground prediction. Second, we measure the predicted foreground region's \emph{intersection over union (IoU)} with the ground-truth foreground region. Finally, we measure the \emph{DICE score}~\cite{dice1945measures} defined as $2|\hat F \cap F| / (|\hat F|+|F|)$,
where $\hat F$ is the predicted foreground region and $F$ is the ground-truth foreground region. In all metrics, higher values mean better performance.

\begin{figure}[!tb]
    \centering
    \begin{minipage}[t]{1.00\linewidth}
    \begin{minipage}[t]{0.19\linewidth}
      \vspace{3pt}
      \centering
      \scriptsize\textbf{\textsf{Image}}
    \end{minipage}
    \hfill
    \begin{minipage}[t]{0.19\linewidth}
      \vspace{3pt}
      \centering
      \scriptsize\textbf{\textsf{Ground-truth}}
    \end{minipage}
    \hfill
    \begin{minipage}[t]{0.19\linewidth}
      \vspace{3pt}
      \centering
      \scriptsize\textbf{\textsf{IEM Result}}
    \end{minipage}
    \hfill
    \begin{minipage}[t]{0.19\linewidth}
      \vspace{0pt}
      \centering
      \scriptsize\textbf{\textsf{Inpainted Image}}
    \end{minipage}
    \hfill
    \begin{minipage}[t]{0.19\linewidth}
      \vspace{0pt}
      \centering
      \scriptsize\textbf{\textsf{IEM+SegNet Result}}
    \end{minipage}
    \end{minipage}
    \includegraphics[width=\linewidth]{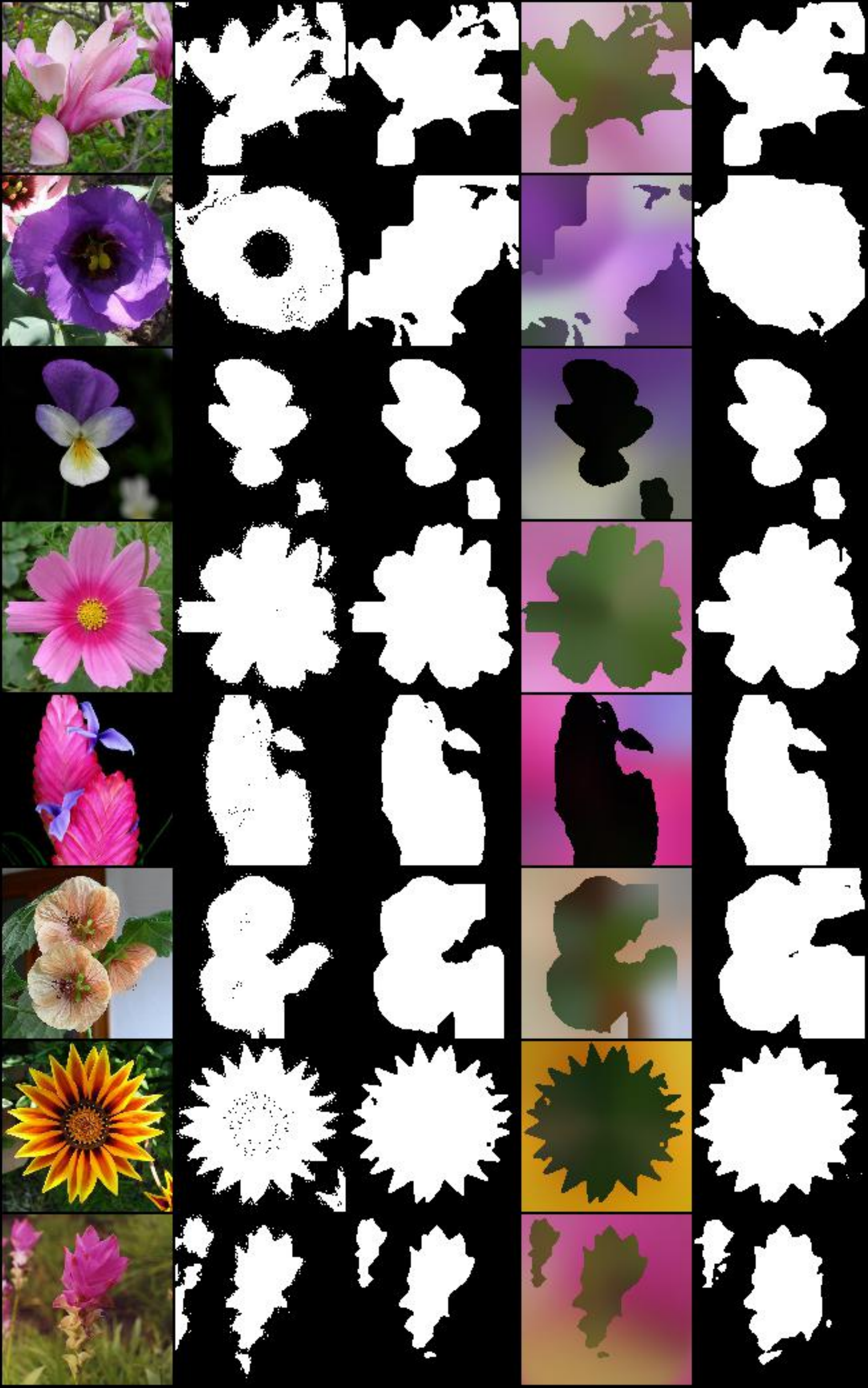}
    \caption{Results of IEM on Flowers. \emph{Left to right:} input image; ground truth mask; IEM mask; inpainting result, with every pixel inpainted as FG or BG according to the IEM mask; SegNet mask} \label{fig:flowers}
    \vspace{-5pt}
\end{figure}

\begin{figure}[!tb]
    \centering
    \begin{minipage}[t]{1.00\linewidth}
    \begin{minipage}[t]{0.19\linewidth}
      \vspace{3pt}
      \centering
      \scriptsize\textbf{\textsf{Image}}
    \end{minipage}
    \hfill
    \begin{minipage}[t]{0.19\linewidth}
      \vspace{3pt}
      \centering
      \scriptsize\textbf{\textsf{Ground-truth}}
    \end{minipage}
    \hfill
    \begin{minipage}[t]{0.19\linewidth}
      \vspace{3pt}
      \centering
      \scriptsize\textbf{\textsf{IEM Result}}
    \end{minipage}
    \hfill
    \begin{minipage}[t]{0.19\linewidth}
      \vspace{0pt}
      \centering
      \scriptsize\textbf{\textsf{Inpainted Image}}
    \end{minipage}
    \hfill
    \begin{minipage}[t]{0.19\linewidth}
      \vspace{0pt}
      \centering
      \scriptsize\textbf{\textsf{IEM+SegNet Result}}
    \end{minipage}
    \end{minipage}
    \includegraphics[width=\linewidth]{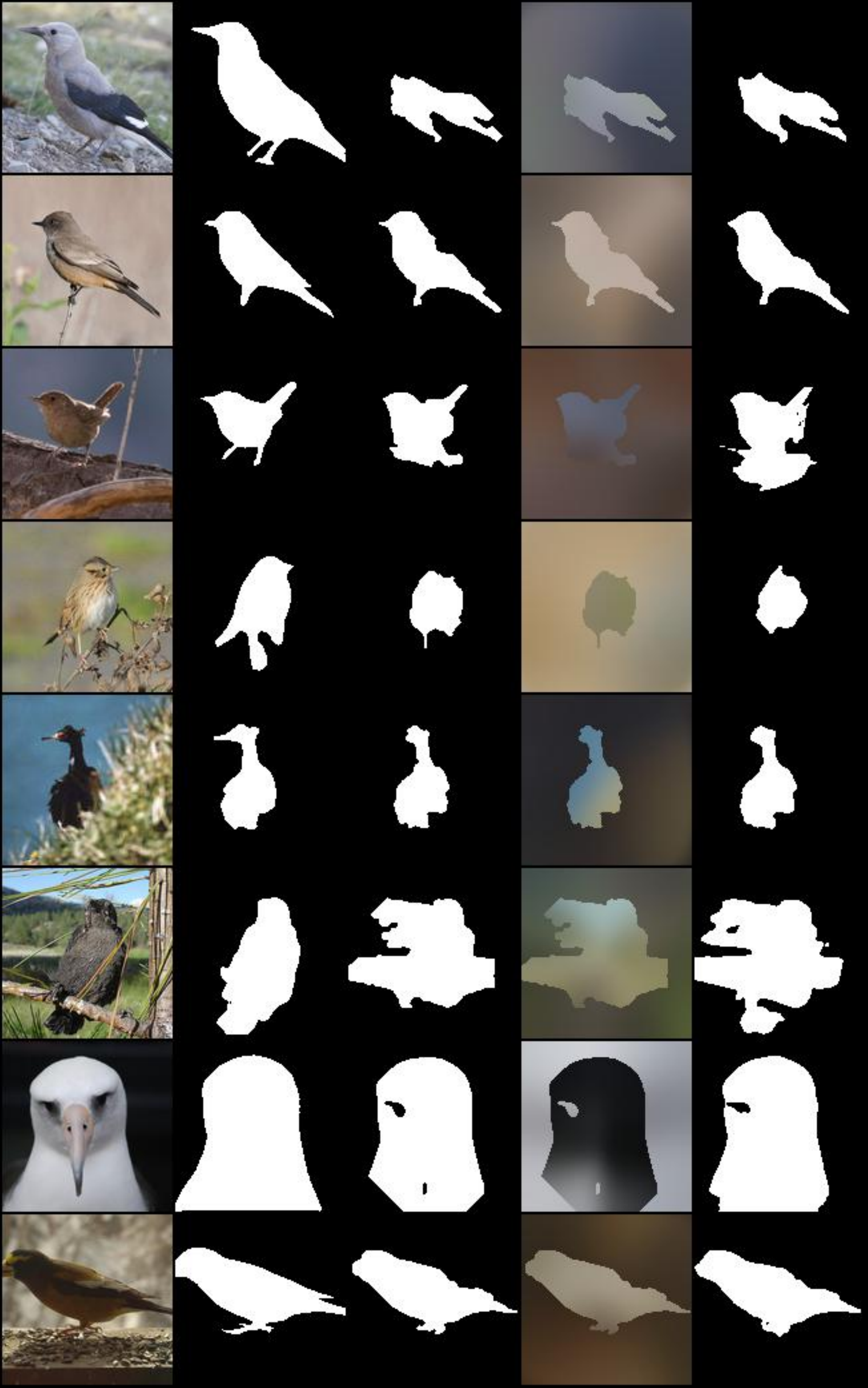}
    \caption{Results of IEM on CUB. \emph{Left to right:} input image; ground truth mask; IEM mask; inpainting result, with every pixel inpainted as FG or BG according to the IEM mask; SegNet mask} \label{fig:birds}
    \vspace{-5pt}
\end{figure}

\begin{figure}[!tb]
    \centering
    \begin{minipage}[t]{1.00\linewidth}
    \begin{minipage}[t]{0.19\linewidth}
      \vspace{3pt}
      \centering
      \scriptsize\textbf{\textsf{Image}}
    \end{minipage}
    \hfill
    \begin{minipage}[t]{0.19\linewidth}
      \vspace{3pt}
      \centering
      \scriptsize\textbf{\textsf{Ground-truth}}
    \end{minipage}
    \hfill
    \begin{minipage}[t]{0.19\linewidth}
      \vspace{3pt}
      \centering
      \scriptsize\textbf{\textsf{IEM Result}}
    \end{minipage}
    \hfill
    \begin{minipage}[t]{0.19\linewidth}
      \vspace{0pt}
      \centering
      \scriptsize\textbf{\textsf{Inpainted Image}}
    \end{minipage}
    \hfill
    \begin{minipage}[t]{0.19\linewidth}
      \vspace{0pt}
      \centering
      \scriptsize\textbf{\textsf{IEM+SegNet Result}}
    \end{minipage}
    \end{minipage}
    \includegraphics[width=\linewidth]{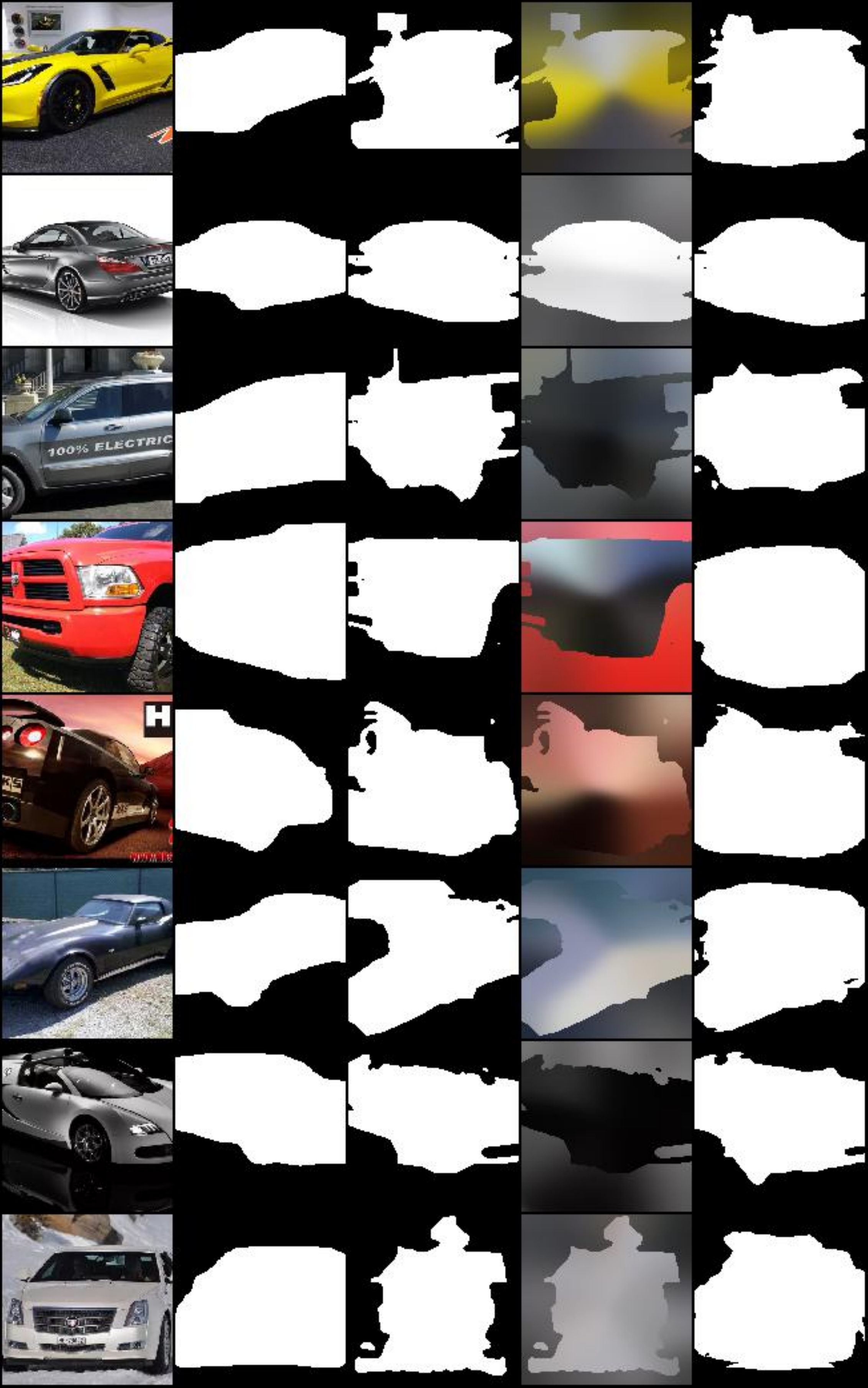}
    \caption{Results of IEM on LSUN Car. \emph{Left to right:} input image; ground truth mask; IEM mask; inpainting result, with every pixel inpainted as FG or BG according to the IEM mask; SegNet mask} \label{fig:cars}
    \vspace{-5pt}
\end{figure}

\subsection{Qualitative Analysis}
In Figures~\ref{fig:flowers},~\ref{fig:birds},~\ref{fig:cars}, we show qualitative results of IEM; all the examples were sampled randomly from the respective test sets, without curation.
In many cases, IEM accurately detects and segments the foreground object. It is successful for many flowers (Figure~\ref{fig:flowers}), where there is often a clear color difference between the foreground (flower) and the background (leaves), but also for many birds even when the background is more complex and the colors of background and foreground are similar.  
Moreover, we observe that the inpainting results are inversely correlated with the mask quality: better masks tend to produce worse inpanting, as expected under IEM. The optional refinement phase with the segmentation model (IEM+SegNet result) tends to improve the masks, albeit by a modest margin.

\subsection{Quantitative Comparison to Prior Work}
We next compare our method to two recent unsupervised segmentation methods, ReDO and PerturbGAN, and to the classic GrabCut algorithm.
In Table~\ref{tab:flowers}, we compare segmentation results on the Flowers dataset. We find that masks from our first phase (IEM) outperform those from ReDO in terms of accuracy and IoU. We see further improvements for masks refined with the segmentation model (IEM+SegNet).
While accuracy and IoU are the only metrics reported for ReDO, we also report the Dice score for completeness and future use. Overall, both the qualitative and the quantitative results suggest that the produced masks are high quality.

In Tables~\ref{tab:birds} and~\ref{tab:cars}, we show results on CUB and LSUN Car. 
The recent GAN-based methods, ReDO and PerturbGAN, outperfom GrabCut, but are likely limited by the known shortcomings of GANs such as mode collapse and unstable training. In contrast, IEM is computationally cheap, class-agnostic, and even applicable to a single unlabeled image.
Consistent with Table~\ref{tab:flowers}, IEM outperforms GrabCut, ReDO, and PerturbGAN, often by a big margin. The refined masks (IEM+SegNet) show consistent further improvement, achieving superior performance across all metrics and datasets.

\subsection{Ablation Studies}

Studies regarding design choices of IEM are discussed in detail in the Appendix and briefly summarized here. Appendix~\ref{app:ablation-training} shows through ablation experiments that different components of IEM are necessary to achieve optimal results, namely regularizing the pixel-wise deviations (Equation~\ref{eq-reg}), smoothing masks after each step, and restricting updates to the mask's boundary.

Moreover, Appendix~\ref{app:ablation-norm} investigates the distributional assumptions adopted for the IEM objective in Equation~\ref{eq-iem}, showing that alternative approximations yield comparable, although inferior, segmentation results on CUB.

Finally, Appendix~\ref{app:ablation-inp} studies the choice of the inpainting component in IEM, where experiments show that, perhaps surprisingly, adopting state-of-the-art inpainters causes considerable degradation of IEM's segmentation performance.

%% file: discussion.tex
\section{Conclusion}
\label{sec:conclusion}

IEM is a simple approach to unsupervised image segmentation that outperforms
competitors reliant on far heavier learning machinery and more computationally
expensive models.  More broadly, our results demonstrate a successful recipe
for unsupervised learning, distinct from both representation learning guided by
proxy tasks, as well as deep learning-based generative modeling approaches.

This recipe calls for optimizing an objective strongly tied to the task of
interest.  Formulation of the optimization problem is actually agnostic to the
use of learning.  For unsupervised segmentation, we found a
criterion motivated by information-theoretic considerations sufficiently
powerful to advance the state-of-the-art.  However, future instantiations of
this recipe, applied to segmentation or other domains, could seek to bring
learning machinery into this first phase.  In either case, solutions to the
optimization objective yield predictions for the target task,
allowing a learning-centric second phase to distill and generalize these
predictions by training a deep neural network to replicate them.

~\\
\noindent\textbf{Acknowledgments}: This work was in part supported by AFOSR Center of Excellence Award, FA9550-18-1-0166.

%% file: appendix.tex
\appendix

\part*{Appendix}

\section{Further Details on IEM Objective}
\label{app:iem-details}

The IEM objective in Equation~\ref{eq-iem} relies on adopting specific $\ell$-norms to approximate the entropy terms in the coefficient of constraint. In particular, we rely on two main approximations, which we describe in detail below.

First, for the conditional entropy of the predicted foreground $\hat F_\phi$ given the predicted background $\hat B_\phi$ (and vice-versa), we have
\begin{equation}
    \begin{split}
     H(\hat F_\phi| \hat B_\phi) &=  H \left(\img\odot \phi(\img) \big| \img \odot \overline{\phi(\img)})\right) \\
        & = - \mathbb E_X \left[  \log P \left(\img \odot \phi(\img) \big| \img \odot \overline{\phi(\img)} \right) \right] \\
        & \approx \mathbb E_X \left[ \left\| \img \odot \phi(\img) - \psi(\img \odot \overline{\phi(\img)}) \right\|_1 \right] \,,
    \end{split}
\end{equation}
where the approximation adopted in the last step amounts to assigning a $\ell_1$-Laplace distribution with identity covariance to the conditional pixel probabilities:
\begin{equation}
    \begin{split}
     P \Big(\img  & \odot \phi(\img) \big| \img \odot \overline{\phi(\img)} \Big) \\
     & = \mathcal L \left(\img \odot \phi(\img) ; \mu \left(\img \odot \overline{\phi(\img)} \right), I \right) \\
     & \propto \exp \left( - \left\| \img \odot \phi(\img) - \mu\left( \img \odot \overline{\phi(\img)} \right)   \right\|_1 \right) \,.
    \end{split}
\end{equation}

Second, for the marginal entropies of the predicted foreground and background, we adopt
\begin{equation}
    \begin{split}
     H(\hat F_\phi) &=  H \left(\img\odot \phi(\img)\right) \\
        & = - \mathbb E_X \left[  \log P \left(\img \odot \phi(\img) \right) \right] \\
        & \approx \mathbb E_X \left[ \left\| \phi(\img) \right\| \right] \,,
    \end{split}
\end{equation}
where $\| \phi(\img)\|$ can be seen as any $\ell^p$ norm: since $\phi(\img)$ is binary, we have that $\|\phi(\img)\|_p = \|\phi(\img)\|_q$ for any $p,q \in [1, \infty)$. Since modelling marginal distributions over images is known to be hard, we opt for an assumption-free approach and assume that pixel values are uniformly distributed, \ie the approximation in the last step above corresponds to the assumption
\begin{equation}
     P\left(\img\odot \phi(\img)\right) = \mathcal U(k)^{\|\phi(\img)\|} = k^{-\|\phi(\img)\|} \,,
\end{equation}
where $k$ captures the number of possible values for a pixel, \eg, $255^3$ for RGB images where each pixel channel is encoded as 8 bits. Note that $\|\phi(\img)\|$ in the equation above represents the number of $1$-valued elements in $\phi(\img)$, hence it can be taken to be any $\ell_p$ norm (or any other function that matches this definition for binary inputs).

\section{Analysis on Training Components of IEM}
\label{app:ablation-training}
To understand the effect of IEM's components, we conduct ablation experiments on CUB and Flowers. We follow the same setup adopted for experiments in Section~\ref{sec:experiments}, running IEM for 150 iterations on the test set of each dataset.

First, we experiment with removing the regularization on foreground and background deviation (Equation~\ref{eq-reg}) by setting $\lambda = 0$ in $\mathcal L_{IEM}$.
Second, we remove the smoothing procedure after mask updates.
Third, we allow mask updates at pixels other than the boundary. 

In Table~\ref{tab:ablation}, we report IoU of produced masks for each experiment. 
Compared to the results with default parameters, mask quality drops in all three ablation experiments, suggesting that these components are important for IEM to achieve the best results. The regularization seems particularly important for Flowers, since it promotes homogeneous colors in the foreground and the background when the images have a clear color contrast between the two. Smoothing masks and limiting updates to the mask boundary seems more important in CUB, where the images have more complex backgrounds, as they prevent the bird segmentations from including other objects (\eg, branches, grass).

\begin{table}[!tb]
\setlength{\tabcolsep}{4.5pt}
\caption{Ablation experiments on CUB and Flowers. Number indicate IoU of masks produced by IEM.} \label{tab:ablation}
\centering
\vspace{1pt}
\begin{tabular}{@{}lrrrr@{}}
\toprule
                                            & CUB   & Flowers     \\ \cmidrule{2-3}
Default parameters                          & 52.2  & 76.8        \\\hline
No regularization on fore/back deviation    & 50.6  & 68.0        \\
No smoothing on projection               & 47.0  & 75.7        \\
Updates not restricted to mask boundary     & 42.8  & 76.6    \\ \bottomrule
\end{tabular}
\end{table}

\section{Analysis on Approximations in IEM}
\label{app:ablation-norm}

\renewcommand{\arraystretch}{1.8}

\begin{table}[!t]
\setlength{\tabcolsep}{4.5pt}
\caption{CUB results with different variants of the proposed IEM objective, each corresponding to different assigned distributions for conditional and marginal pixel distributions.}
\label{tab:ablation-norm}
\centering
\vspace{1pt}
\begin{tabular}{@{}lrr@{}}
\toprule
Objective (first term)                       & IoU  & DICE  \\ \cmidrule{2-3}
$\frac{\| M \odot( \img - \psi_K(\img \odot \overline M, \overline M)) \|_1}{\|M\|}$ (Equation~\ref{eq-iem-direct}) & \textbf{52.2}  & \textbf{66.0}  \\
$\frac{\| M \odot( \img - \psi_K(\img \odot \overline M, \overline M)) \|_2}{\|M\|}$ (Assumption 1)& 51.7  & 65.6  \\
$\frac{\| M \odot( \img - \psi_K(\img \odot \overline M, \overline M)) \|_1}{\|\img \odot M\|_1}$ (Assumption 3) & 51.8  & 65.6  \\
$\frac{\| M \odot( \img - \psi_K(\img \odot \overline M, \overline M)) \|_2}{\|X \odot M\|_2}$ (Assumptions 1+2)    & 51.2  & 65.1  \\
\bottomrule
\end{tabular}
\end{table}

As discussed in Appendix~\ref{app:iem-details}, our proposed IEM objective adopts two key approximations for the conditional and marginal entropies in the original coefficient of constraint minimization problem in Equation~\ref{eq-mincoef}. Although the Laplacian approximation for conditional pixel probabilities is popular in the computer vision literature, for example in papers on inpainting \cite{inpaintinggated, inpaintingattention} and image modelling \cite{pix2pix2017, CycleGAN2017}, it is unclear whether it is the optimal choice for our setting.

Additionally, the uniform prior over pixel values that we adopt to approximate marginal entropies can be seen as being overly simple, especially since different priors are more commonly adopted in the literature \eg, zero-mean isotropic Gaussians.

To investigate whether our approximations are sensible, we consider three variants of the proposed IEM objective, each being the result of different approximations for the image entropies. In particular, we consider:
\begin{enumerate}
    \item Assuming that the conditional pixel probabilities follow a isotropic Gaussian (instead of a $\ell_1$-Laplacian), which yields the approximation
    \begin{equation}
        \begin{split}
        H(& \hat F_\phi| \hat B_\phi) \\
        & \approx \mathbb E \left[ \left\| \img \odot \mu(\img) - \psi(\img \odot \overline{\phi(\img)}) \right\|_2 \right] \,,
        \end{split}
    \end{equation}
    which in practice amounts to adopting the $\ell_2$ norm instead of $\ell_1$ in the numerators.
    \item Assuming that the marginal foreground/background distributions are zero-mean isotropic Gaussians, which results in
    \begin{equation}
        H(\hat F_\phi) \approx \mathbb E \left[ \left\| \img \odot \phi(\img) \right\|_2 \right] \,.
    \end{equation}
    \item Assuming that the marginal foreground/background distributions are zero-mean $\ell_1$-Laplacians with identity covariance, yielding
    \begin{equation}
        H(\hat F_\phi) \approx \mathbb E \left[ \left\| \img \odot \phi(\img) \right\|_1 \right] \,.
    \end{equation}
\end{enumerate}

We repeat our experiments on the CUB dataset, following the same protocol described in Section~\ref{sec:experiments}, \ie masks are optimized for a total of 150 iterations to maximize the corresponding objective, and $\psi_K$ is the same fixed inpainter as in our original experiments.

Table~\ref{tab:ablation-norm} summarizes our results, showing that although our chosen approximations yield the best segmentation performance measured in IoU and DICE score, all variants of the IEM objective offer comparable results. This suggests that our proposed framework does not strongly rely on our particular distributional assumptions (or, equivalently, to the adopted norms for the inpainting objective), offering a general approach for unsupervised segmentation.

\section{Analysis on Inpainting Component}
\label{app:ablation-inp}

The inpainter we adopted for all experiments in Section~\ref{sec:experiments} is significantly simpler than inpainting modules typically employed in other works, consisting of a single $21 \times 21$ convolution with a Gaussian filter. Such module has the advantage of having a small computational cost and not requiring any training, making it suitable for a learning-free method.

Here, we show that such simple inpainting module also yields better segmentation masks when compared to more sophisticated variants. Table~\ref{tab:ablation-inpainter} shows the quality of masks produced by IEM when adopting the inpainting component proposed in Yu~\etal~\cite{inpaintinggated}, which consists of gated convolutions and contextual attention, and is trained with the $\ell_1$ loss along with an adversarial objective produced by a patch-wise discriminator (`GatedConv' entry in the table).

`GatedConv (coarse outputs)' refers to IEM results when taking the coarse outputs of GatedConv to compute the IEM objective: more specifically, we take the `GatedConv' model (trained with both the $\ell_1$ and adversarial loss) but only pass the foreground/background image through the first half of the network, which generates a coarse inpainted image that precedes the contextual attention layers (see Figure 3 of Yu~\etal~\cite{inpaintinggated} for reference).  `GatedConv ($\ell_1$ only)' refers to the GatedConv model trained only with the $\ell_1$ loss (\ie without SN-PatchGAN), with coarse outputs only. 
All GatedConv models were pre-trained on the whole CUB dataset with free-form masks \cite{inpaintinggated} and then held fixed during IEM. When evaluated in terms of IoU and DICE, the quality of masks produced by IEM deteriorates monotonically with the complexity of the inpainting component: the original GatedConv model yields the lowest segmentation scores, which improves if IEM is run against its intermediate, coarse inpaintings, and training the network without contextual attention or the adversarial loss yields the best segmentation results other than ours.

We also evaluate how fine-tuning the inpainter during IEM, \ie optimizing the inpainter with masks currently produced by IEM, affects the quality of segmentation masks. Table~\ref{tab:ablation-inpainter} shows that it also deteriorates the quality of masks produced by IEM (compare last two rows).

\renewcommand{\arraystretch}{1.2}

\begin{table}[!t]
\setlength{\tabcolsep}{8.pt}
\caption{Comparison between our simple inpainter and variants of the Gated Convolutional (GatedConv) model proposed in Yu~\etal~\cite{inpaintinggated}, in term of quality of masks produced on CUB. Removing components from GatedConv, such as removing its refinement phase during IEM (`GatedConv, coarse outputs') and training without adversarial losses (`GatedConv, $\ell_1$ only') deteriorates its inpainting quality but results in better IEM segmentations.}
\label{tab:ablation-inpainter}
\centering
\vspace{1pt}
\begin{tabular}{@{}lrr@{}}
\toprule
Inpainting Module                     & IoU  & DICE  \\ \cmidrule{2-3}
Simple (Equation~\ref{eq-inpainter})                        & \textbf{52.2}  & \textbf{66.0}  \\
GatedConv \cite{inpaintinggated}                  & 40.3  & 55.8  \\
GatedConv, coarse outputs \cite{inpaintinggated}                   & 41.6  & 56.8  \\
GatedConv, $\ell_1$ only \cite{inpaintinggated} & 43.7  & 59.0  \\
GatedConv+Fine tuning, $\ell_1$ only \cite{inpaintinggated}    & 41.7  & 57.1  \\
\bottomrule
\end{tabular}
\end{table}